\documentclass[10pt]{article}
\usepackage[margin=0.9in]{geometry}
\usepackage{amsmath,amssymb,amsfonts}
\usepackage{booktabs}
\usepackage{enumitem}
\usepackage{graphicx}
\graphicspath{{./}{doc/}}
\usepackage{hyperref}
\hypersetup{hidelinks}
\usepackage{microtype}
\usepackage{indentfirst}
\usepackage{mathtools}
\usepackage{bm}
\usepackage{tabularx}
\usepackage{tikz}
\usepackage{float}
\usetikzlibrary{positioning,arrows.meta,fit,calc}

\setlist{nosep,leftmargin=*}

\newcommand{\KTCM}{K_{\mathrm{TCM}}}
\newcommand{\Normal}{\mathcal{N}}
\newcommand{\KL}{\mathrm{KL}}

\providecommand{\IEEEauthorblockN}[1]{#1\\[0.5ex]}
\providecommand{\IEEEauthorblockA}[1]{\begin{tabular}[t]{c}#1\end{tabular}}

\title{Alpha-TCM: A Nine-Layer Framework for TCM-Guided\\Multimodal Inference and Structured-Prior Auditing}
\author{\IEEEauthorblockN{Yucheng Liu\textsuperscript{1}\quad Everett X. Wang\textsuperscript{2,*}}
\IEEEauthorblockA{\textsuperscript{1}Department of Engineering Science, University of Oxford, Oxford, United Kingdom\\
\textsuperscript{2}Beijing Institute of Nanoenergy and Nanosystems, Chinese Academy of Sciences, Beijing, China\\
\textsuperscript{*}Corresponding author: Everett X. Wang
(\href{mailto:wangxiaolin@binn.cas.cn}{wangxiaolin@binn.cas.cn})}
}
\date{}

\begin{document}
\maketitle

\begin{abstract}
Alpha-TCM makes two co-primary contributions under one thesis: auditable systems should keep domain knowledge, case evidence, probabilistic inference, and downstream authority distinct. First, it establishes a nine-layer TCM-oriented architecture that separates semantic definitions, multimodal acquisition, validity and quality, provenance-bearing knowledge, probabilistic state inference, longitudinal updating, constrained decisions, language interaction, and human governance. Each layer is defined by an information contract, failure boundary, downstream dependency, and validation status; the architecture is a system-design contribution rather than a claim that all nine layers are clinically implemented. Second, the paper instantiates the static Layers 1--5 pathway in a controlled synthetic prototype and audits the framework's central knowledge--evidence interface. Four conditional-Gaussian encoders process tongue-like images, pulse-like waveforms, masked inquiry fields, and locally generated text, while a frozen bridge predicts a continuous five-dimensional synthetic outcome. A source-registered, researcher-specified synthetic contextual prior is compiled from versioned knowledge files; a separately implemented renderer never imports the compiler or its parameter object. Conditional-expert fusion removes duplicated reference-prior information before introducing the contextual prior. Equal-information controls, encoder-level prior regularization, graded contamination, and structural corruptions distinguish contextual benefit from specifically Bayesian benefit and predictive harm from conflict detection. In a create-once confirmation over ten paired seeds and four training sizes, contextual knowledge improved MAE over marginal committee fusion at 600 and 1,200 cases, but did not outperform equal-information neural controls. Fully wrong context caused harm at every size, whereas encoder-level prior effects were inconsistent and the conflict score did not track case-level damage. Thus the mixed audit supports useful low-data contextual structure, not a specifically Bayesian predictive advantage. The prototype tests only this hypothetical synthetic operationalization; it cannot validate or invalidate TCM theory or support clinical, diagnostic, treatment, causal, longitudinal, or transfer claims.
\end{abstract}

\section{Introduction}

Traditional Chinese medicine (TCM) assessment is commonly described as multimodal and sequential: inspection, auscultation and olfaction, inquiry, and palpation are interpreted together with history and change over time \cite{zhang2021survey}. From a machine-learning perspective, this is not merely a larger classification problem. The observations have different data types, may be missing, and may vary in acquisition quality; practitioner interpretations can disagree \cite{jacobson2019agreement}; and the state of interest is only partially observed. A system that produces one confident label without representing these distinctions can conflate absence with negative evidence, measurement failure with patient state, and language generation with inference.

Computational TCM research has produced useful tongue-image models, pulse-acquisition protocols, structured terminology, and more recent multimodal systems \cite{jiang2022tongue,chang2024tongue,zhu2023pulse,zhang2025ccl,luo2026mmir}. These efforts demonstrate that individual parts of the workflow can be digitized. They do not by themselves specify how structured domain knowledge, heterogeneous evidence, uncertainty, repeated observations, and downstream human review should interact. Alpha-TCM addresses that integration problem at the level of variables, probabilistic interfaces, and failure boundaries rather than proposing that every layer requires a new neural architecture.

Let $Z_t$ denote a continuous latent state at visit $t$, $X_t=\{X_t^{(m)}\}$ a set of modality observations, $M_t$ acquisition and quality metadata, $Y_t^{(r)}$ an assessment by practitioner $r$, and $\KTCM$ a provenance-bearing representation of domain knowledge. The general inferential object is
\begin{equation}
p\!\left(Z_t\mid X_t,M_t,Y_t^{(1:R)},\KTCM\right),
\label{eq:posterior-object}
\end{equation}
not a deterministic disease or treatment label. TCM concepts enter as operational constructs and potentially revisable prior structure. Their usefulness would need to be tested by reproducibility, calibrated prediction, and external validation rather than assumed from the notation.

The individual ingredients---neural modality encoders, Gaussian predictive heads, committee fusion, mixture priors, and continuous downstream prediction---are not claimed as separate algorithmic inventions. The paper instead makes two contributions at equal conceptual strength but with different evidence types. The first is a system architecture: Alpha-TCM defines where semantics, acquisition, quality, knowledge, inference, time, decisions, language, and governance belong, what each layer may consume and emit, and which failures must not cross layer boundaries. The second is an empirical prototype: a static vertical slice instantiates Layers 1--5 and tests whether a provenance-bearing TCM operationalization supplies useful contextual inductive structure under matched controls.

The prototype is therefore not an unrelated fusion benchmark. It tests one consequence of the paper's unifying separation of knowledge, evidence, inference, and authority: the framework's central knowledge--evidence contract. Domain theory is assigned a revisable prior role; heterogeneous observations contribute uncertain conditional evidence; marginal prior information is not counted repeatedly; and absent, learned, aligned, contaminated, and structurally misspecified controls make the operationalized prior capable of failing. Predictive performance, uncertainty, wrong-prior damage, and conflict feedback are reported separately because success on one does not imply success on the others.

The framework remains broader than the evaluated path. The evaluated prototype implements four modality-specific conditional Gaussian encoders, marginal and context-conditioned committee fusion, equal-information neural controls, a continuous probabilistic bridge, encoder-level distributional regularization, structural corruptions, and a descriptive conflict audit. It does \emph{not} implement learned validity or quality heads, practitioner modeling, longitudinal forecasting, decision optimization, LLM inferential authority, or clinical governance. Those layers are specified as system contracts and validation requirements, not presented as completed empirical components.

The contributions are:
\begin{itemize}
  \item \textbf{Nine-layer system architecture:} a TCM-oriented separation of semantics, acquisition, validity and quality, structured knowledge, probabilistic inference, longitudinal modeling, constrained decisions, language interaction, and human governance, with explicit contracts, failure boundaries, implementation status, and validation requirements;
  \item \textbf{Formal knowledge--evidence interface:} an auditable route from versioned TCM assertions to a challengeable contextual prior, together with conditional-posterior accounting that removes repeated reference-prior information before inserting $p_{\KTCM}(Z\mid C)$;
  \item \textbf{Prototype instantiation:} a reproducible static vertical slice of Layers 1--5 with four heterogeneous encoders, uncertainty propagation, a continuous bridge, and equal-information neural controls; and
  \item \textbf{Falsifiable prior audit:} a four-size, ten-seed, evidence-gated experiment with unconditional, aligned, wrong-context, contaminated, and structurally corrupted priors, followed by one-time confirmation and conservative claim rules.
\end{itemize}

\section{Related Work}

\subsection{Computational TCM and Structured Knowledge}

Reviews of computational TCM describe work on syndrome determination, disease prediction, prescription analysis, and digitization of the traditional examinations \cite{chen2022mltcm,zhang2021survey}. Tongue analysis is among the most developed image-based tasks. Jiang et al. trained a multi-label tongue-image detector, Chang et al. constructed a manually annotated feature dataset for real-time recognition, and Gao et al. reported a broad tongue-model ablation study \cite{jiang2022tongue,chang2024tongue,gao2026tongue}. Recent studies have combined image and text components or connected language models to multi-stage diagnosis-and-treatment tasks \cite{zhang2025ccl,luo2026mmir}. Alpha-TCM is complementary: its focus is not another end-to-end classifier or prescription generator, but an auditable probabilistic interface between observations, structured knowledge, uncertain state estimates, and later human-supervised use.

Standard terminology and ontology research shows that TCM concepts can be represented consistently enough for information exchange, but also documents ambiguity and maintenance challenges \cite{who2007terminology,long2019ontology}. We therefore treat ontology relations as sourced computational assertions with declared epistemic status, not as direct biological facts. This view resembles work that combines medical ontologies with learned patient representations, including MMUGL and KEEP \cite{burger2023mmugl,elhussein2025keep}. Alpha-TCM expresses the same general principle in Bayesian form: structured knowledge may shape a prior, while observed evidence may weaken or contradict it.

\subsection{Multimodal Fusion, Conditional Experts, and Prior Conflict}

Clinical multimodal systems must handle heterogeneous types, incomplete modality combinations, and acquisition-dependent reliability. MedPatch and HEALNet illustrate confidence-guided or flexible fusion under heterogeneous and missing modalities \cite{jorf2025medpatch,hemker2024healnet}. Multimodal latent-variable work includes PoE inference in MVAE, MoE inference in MMVAE, and mixture-of-products aggregation in MoPoE-VAE \cite{wu2018mvae,shi2019mmvae,sutter2021mopoe}. Those generative models specify how modality likelihoods and a common prior induce subset posteriors. Our encoders instead learn conditional Gaussian approximations discriminatively, so their product requires an explicit repeated-prior correction.

The Bayesian committee machine combines conditional estimators by dividing out duplicated prior information; robust and generalized variants additionally temper expert contributions \cite{tresp2000bcm,deisenroth2015rbcm}. Products of Gaussian experts can otherwise become overconfident or behave erratically, motivating explicit calibration and diagnostics \cite{cohen2020healing}. The evaluated prototype therefore uses marginal committee correction as its noncontextual probabilistic reference and compares contextual knowledge with equal-information learned-context and late-concatenation controls. Applying the accounting rule to this prototype is part of the formal interface, not a claim of a new committee-machine theorem.

Prior--data conflict is ordinarily assessed relative to a prior-predictive distribution, not by an arbitrary distance threshold \cite{evans2006conflict}. Our earlier nearest-component flag is therefore retained only as a descriptive geometry score after training-only width calibration. It is not a Bayesian conflict test and cannot adjudicate a medical theory. Neural predictive uncertainty can also be poorly calibrated under shift \cite{guo2017calibration,ovadia2019uncertainty}; Bayesian notation alone is not evidence of trustworthy intervals.

\subsection{Longitudinal Modeling and Human Oversight}

Irregularly sampled clinical data motivate models that distinguish latent dynamics from discrete noisy observations. Neural continuous-discrete state-space models and irregular multimodal electronic-health-record models are relevant examples \cite{ansari2023ncdssm,zhang2023irregular}. Alpha-TCM adopts the conceptual separation between state transitions and observation likelihoods, while treating interventions as predictive covariates unless a separate causal design is justified. The decision and interaction layers similarly follow the principle that uncertainty, escalation, and human authority must be explicit \cite{who2021ai,roy2026oversight}. These longitudinal and governance layers are architectural scope in this paper, not evaluated components.

Established health-AI guidance covers broader and partly overlapping concerns. WHO provides ethical and governance principles, FUTURE-AI organizes lifecycle requirements for trustworthy and deployable systems, and DECIDE-AI specifies reporting for early live clinical evaluation \cite{who2021ai,lekadir2025futureai,vasey2022decideai}. Alpha-TCM does not claim to replace or outperform these consensus and reporting frameworks. Its narrower contribution is a domain-oriented allocation of explicit information contracts and failure boundaries, coupled to an executable audit of one knowledge--evidence interface. Whether this nine-layer allocation is more usable or failure-sensitive than alternative decompositions remains an open empirical question.

\section{System Requirements and Design Principles}
\label{sec:requirements}

Alpha-TCM is organized around six design principles. First, named TCM constructs are operational definitions rather than asserted physical quantities. Second, latent state is distinct from syndrome, disease, or treatment labels. Third, availability, domain validity, measurement quality, and a valid negative observation are different variables. Fourth, practitioner assessments are heterogeneous observations rather than a single unquestioned ground truth. Fifth, longitudinal association is not a causal treatment effect. Sixth, numerical inference, constrained decisions, natural-language communication, and human authority remain separate.

These principles induce system-level requirements rather than a sequence of interchangeable neural modules. Table~\ref{tab:system-requirements} states the invariants that every implementation must preserve. The architecture contribution is the explicit allocation of epistemic responsibilities: later layers may consume declared outputs from earlier layers, but they may not silently redefine their meaning or authority.

\begin{table}[H]
\centering
\small
\begin{tabularx}{\textwidth}{@{}p{0.07\textwidth}p{0.25\textwidth}X p{0.27\textwidth}@{}}
\toprule
Req. & System invariant & Failure prevented & Verification obligation \\
\midrule
R1 & Every construct has an operational definition, source, version, and epistemic status & Terminology treated as biological fact & Semantic and provenance audit \\
R2 & Observation, availability, validity, quality, and negative evidence remain distinct & Missing or invalid input treated as patient state & Acquisition and missingness tests \\
R3 & Knowledge and case evidence use separate interfaces & Domain theory copied into observations or labels & Source and data-flow traceability \\
R4 & Posterior state, downstream prediction, decision, and explanation remain distinct & Prediction silently converted into diagnosis or action & Interface and authority tests \\
R5 & Prior influence is measurable, challengeable, and removable & Structured knowledge becomes unfalsifiable authority & Absent, aligned, and wrong-prior controls \\
R6 & Temporal conditioning and causal claims require separate assumptions & Association presented as treatment effect & Longitudinal and causal-design audit \\
\bottomrule
\end{tabularx}
\caption{System requirements that remain binding across all nine Alpha-TCM layers. They are architectural invariants, not claims of completed clinical validation.}
\label{tab:system-requirements}
\end{table}

Three cross-layer rules follow. First, provenance travels with every semantic assertion, prior, model artifact, and downstream report. Second, uncertainty may be transformed but cannot be silently discarded when crossing an interface. Third, authority narrows downstream: an acquisition layer can record an observation, an inference layer can estimate a state, and a language layer can explain a fixed record, but none thereby gains authority to diagnose, prescribe, or override human review.

\section{Nine-Layer Alpha-TCM Architecture}
\label{sec:architecture}

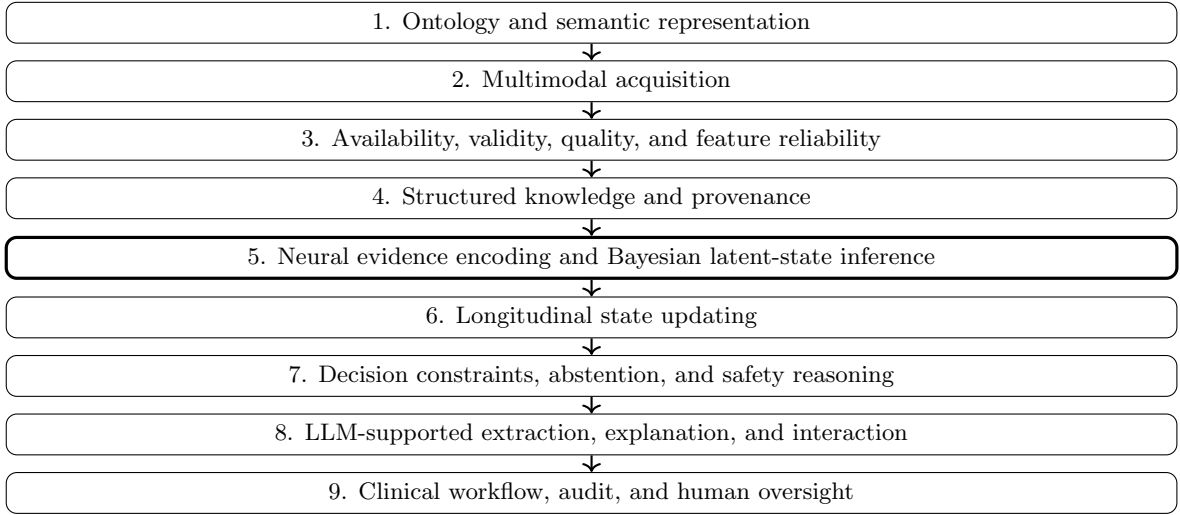
\begin{figure}[H]
\centering
\begin{tikzpicture}[
  node distance=0.22cm,
  layer/.style={draw,rounded corners,minimum width=0.91\textwidth,minimum height=0.54cm,align=center,font=\small},
  arr/.style={->,thick}
]
\node[layer] (l1) {1. Ontology and semantic representation};
\node[layer,below=of l1] (l2) {2. Multimodal acquisition};
\node[layer,below=of l2] (l3) {3. Availability, validity, quality, and feature reliability};
\node[layer,below=of l3] (l4) {4. Structured knowledge and provenance};
\node[layer,below=of l4,very thick] (l5) {5. Neural evidence encoding and Bayesian latent-state inference};
\node[layer,below=of l5] (l6) {6. Longitudinal state updating};
\node[layer,below=of l6] (l7) {7. Decision constraints, abstention, and safety reasoning};
\node[layer,below=of l7] (l8) {8. LLM-supported extraction, explanation, and interaction};
\node[layer,below=of l8] (l9) {9. Clinical workflow, audit, and human oversight};
\foreach \a/\b in {l1/l2,l2/l3,l3/l4,l4/l5,l5/l6,l6/l7,l7/l8,l8/l9}{\draw[arr] (\a) -- (\b);}
\end{tikzpicture}
\caption{Alpha-TCM's nine-layer dependency order. The architecture covers the complete information pathway; the evaluated prototype instantiates only the static Layers 1--5 vertical slice.}
\label{fig:nine-layer}
\end{figure}

Figure~\ref{fig:nine-layer} and Table~\ref{tab:nine-layer} summarize the architecture. The layers are dependency and failure-containment boundaries rather than nine claims of algorithmic novelty. Layers 1--4 define meaning, observations, quality, and knowledge; Layer 5 is the static probabilistic core; Layer 6 introduces temporal state; and Layers 7--9 govern downstream action, communication, and use.

\begin{table}[H]
\centering
\small
\begin{tabularx}{\textwidth}{@{}p{0.05\textwidth}p{0.28\textwidth}X p{0.21\textwidth}@{}}
\toprule
Layer & Principal contract & Failure contained & Evaluated status \\
\midrule
1 & Versioned identifiers, definitions, sources, and epistemic status & Semantic ambiguity and category collapse & Synthetic contract implemented \\
2 & Raw observations plus acquisition metadata & Modality and acquisition conflation & Four synthetic modalities \\
3 & Separate availability, validity, quality, and negative findings & Missingness treated as evidence & Inquiry mask only \\
4 & Versioned, challengeable relations and prior potentials & Theory treated as unchangeable fact & Contextual prior compiled \\
5 & Conditional evidence, prior accounting, posterior state, uncertainty & Overconfident or duplicated fusion & Encoders, BCM, contextual fusion \\
6 & Transition, filtering, smoothing, and forecasting contracts & Visits treated as independent snapshots & Specified, not evaluated \\
7 & Utility, constraints, abstention, and information seeking & Prediction treated as authorized action & Specified, not evaluated \\
8 & Extraction and explanation without numerical authority & Language generation overriding inference & Text input only \\
9 & Human review, logging, access control, and monitoring & Unaccountable deployment & Specified, not evaluated \\
\bottomrule
\end{tabularx}
\caption{Nine-layer contracts, the failures they contain, and the exact boundary of the current prototype.}
\label{tab:nine-layer}
\end{table}

\subsection{Layers 1--4: Meaning, Observation, Quality, and Knowledge}

Layer 1 requires each computational concept to have an identifier, operational definition, source, version, and declared evidence type. Terminological, TCM-theoretical, and biomedical statements must not be silently merged. Layer 2 records an observation together with acquisition context. Layer 3 distinguishes availability $R_t^{(m)}$, domain validity $V_t^{(m)}$, measurement quality $\rho_t^{(m)}$, and a valid negative finding. Layer 4 stores domain relations as challengeable structure from which a prior may be compiled. The evaluated prototype implements a narrow Layer-4 route through a synthetic pattern taxonomy, latent-mapping hypotheses, source registry, and contextual-prior contract. It does not implement learned $V_t^{(m)}$ or $\rho_t^{(m)}$ heads.

\subsection{Layer 5: Probabilistic State Inference}

Layer 5 transforms heterogeneous observations into distributions over a shared state and combines them without erasing uncertainty or modality availability. The prototype uses marginal BCM to remove duplicated reference-prior information, contextual fusion to insert the versioned knowledge prior, and encoder-level KL regularization to test whether aligned or wrong privileged structure changes representation learning. Equal-information learned-context and late-concatenation controls receive the same observed context features. This layer returns a state distribution and a downstream synthetic prediction, not a diagnosis or treatment decision.

\subsection{Layers 6--9: Time, Decisions, Language, and Governance}

Layer 6 defines state transitions rather than treating visits as independent. Layer 7 maps posterior information to constrained actions such as abstention or requesting additional evidence. Layer 8 may extract structured observations or verbalize a fixed numerical record, but it cannot silently replace the posterior. Layer 9 retains human authority, audit trails, access control, and monitoring. These layers are substantive system contracts and validation targets, but synthetic Layer-5 performance is not evidence that they are clinically valid or safe.

\section{Prototype Instantiation of Layers 1--5}
\label{sec:prototype}

The evaluated prototype is a vertical slice designed to test the most central cross-layer hypothesis: whether versioned TCM knowledge can influence uncertain multimodal inference through an explicit, falsifiable interface. Table~\ref{tab:prototype-trace} maps the architecture to executable artifacts. This traceability prevents a successful Layer-5 experiment from being presented as validation of the complete system while also showing that the prototype directly instantiates, rather than merely illustrates, the architecture.

\begin{table}[H]
\centering
\small
\begin{tabularx}{\textwidth}{@{}p{0.10\textwidth}p{0.34\textwidth}X@{}}
\toprule
Layer & Prototype artifact & Tested property \\
\midrule
1 & Versioned synthetic taxonomy, latent-mapping hypotheses, and source registry & Definitions and provenance are explicit and hashable \\
2 & Tongue-like image, pulse-like waveform, masked inquiry, text, and context metadata & Heterogeneous observations share a case without sharing an encoder \\
3 & Inquiry observation mask and modality omission interface & Missing information is represented separately from observed values \\
4 & Contextual-prior contract and deterministic knowledge compiler & Knowledge is separate from renderer code and can be corrupted or removed \\
5 & Four conditional experts, marginal and contextual fusion, bridge, uncertainty, and conflict audit & Prior influence, prediction, uncertainty, and disagreement are separately measurable \\
\bottomrule
\end{tabularx}
\caption{Traceability from the nine-layer architecture to the implemented static prototype.}
\label{tab:prototype-trace}
\end{table}

The prototype evaluates three architecture-level propositions. First, knowledge and observations can be implemented through separate provenance-bearing pathways. Second, contextual knowledge can enter probabilistic inference without being copied into modality features or deterministic labels. Third, prior benefit and prior failure can be evaluated under matched absent, learned, aligned, and misspecified conditions. The experiment does not test the validity of TCM theory itself; it tests whether one declared operationalization behaves as useful and revisable system knowledge under a frozen synthetic protocol.

\section{Probabilistic Formulation}
\label{sec:probabilistic}

\subsection{Static Factorization}

For a single visit, a general factorization is
\begin{align}
p&(Z,X,M,Y^{(1:R)},S\mid\KTCM) \nonumber\\
&=p(Z\mid\KTCM)
\prod_m p(X^{(m)}\mid Z,M^{(m)})
\prod_r p(Y^{(r)}\mid Z,\theta_r)
p(S\mid Z,\KTCM),
\label{eq:static-factorization}
\end{align}
where $S$ denotes an optional syndrome or prototype projection and $\theta_r$ practitioner-specific parameters. This separation is compatible with expertise-aware and crowdsourced-annotation models that retain rater heterogeneity instead of collapsing it into one label \cite{yan2010annotator,xia2024crowd}. This factorization assigns distinct roles to prior knowledge, measurements, raters, and interpretation. It does not imply that every factor is implemented in the current experiment.

A structured knowledge prior can be represented as
\begin{equation}
p(Z\mid\KTCM,\lambda)\propto p_0(Z)
\prod_{e\in E_{\mathrm{TCM}}}\psi_e(Z;\lambda_e),
\label{eq:knowledge-prior}
\end{equation}
where $p_0$ is a generic prior, $\psi_e$ is a potential associated with a sourced relation, and $\lambda_e\geq0$ controls its strength. A defensible study compares an asserted structure with absent, learned, and deliberately misspecified controls under matched information and capacity. The prototype instantiates a restricted form through a provenance-bearing synthetic TCM taxonomy and contextual-prior contract compiled independently of the renderer implementation. The sources motivate terminology and declared relations; they do not establish clinical or biomedical validity, and the synthetic test cannot validate or invalidate TCM theory.

\subsection{Conditional Gaussian Encoders and Fusion}

Each implemented encoder is trained discriminatively and therefore approximates a conditional distribution, not a likelihood:
\begin{equation}
q_m(Z\mid X^{(m)})=
\Normal\!\left(Z;\mu_m,\operatorname{diag}(\sigma_m^2)\right),
\qquad \sigma_{mj}^2>0.
\label{eq:modality-evidence}
\end{equation}
The Gaussian NLL used to train an encoder on $Z^{\mathrm{true}}$ is, up to constants,
\begin{equation}
\mathcal{L}_{m}=\frac{1}{2}\sum_j
\left[\log\sigma_{mj}^2+
\frac{(z_j^{\mathrm{true}}-\mu_{mj})^2}{\sigma_{mj}^2}\right].
\label{eq:encoder-nll}
\end{equation}

If every $q_m(Z\mid X^{(m)})$ contains the same population prior $p_0(Z)$ and modalities are conditionally independent given $Z$, multiplying the conditionals counts that prior $M$ times. The corrected committee approximation is
\begin{equation}
q_{\mathrm{BCM}}(Z\mid X^{(\mathcal A)})\propto
p_0(Z)^{1-|\mathcal A|}\prod_{m\in\mathcal A}q_m(Z\mid X^{(m)}).
\label{eq:bcm}
\end{equation}
For diagonal Gaussian $p_0=\Normal(\mu_0,\operatorname{diag}(\sigma_0^2))$, the coordinate-wise natural parameters are
\begin{align}
\tau_{\mathrm{BCM},j}&=\sum_{m\in\mathcal A}\tau_{mj}-(|\mathcal A|-1)\tau_{0j},\nonumber\\
\mu_{\mathrm{BCM},j}&=\tau_{\mathrm{BCM},j}^{-1}
\left[\sum_{m\in\mathcal A}\tau_{mj}\mu_{mj}-(|\mathcal A|-1)\tau_{0j}\mu_{0j}\right],
\label{eq:bcm-gaussian}
\end{align}
where $\tau=\sigma^{-2}$. The reference Gaussian is estimated from training latent targets only. Nonpositive corrected precision means the Gaussian committee approximation is improper; the implementation reports its frequency and applies a declared $10^{-4}$ floor only to keep the bridge numerically defined. No precision-floor event occurred in the frozen confirmation outputs. The uncorrected product
\begin{equation}
q_{\mathrm{PoE}}(Z\mid X^{(\mathcal A)})\propto\prod_{m\in\mathcal A}q_m(Z\mid X^{(m)})
\label{eq:poe}
\end{equation}
is retained as an explicitly non-Bayesian ablation. Missing modalities are omitted rather than fabricated.

\subsection{Source-Registered, Researcher-Specified Synthetic Contextual Prior}

The implemented knowledge prior is compiled from four versioned inputs: a synthetic TCM pattern taxonomy, declared latent-mapping hypotheses, a source registry, and a contextual-prior contract. The external sources support terminology and historical provenance; they do not establish the numerical latent mappings or diagnostic validity. Cue vectors, component radius, axis variances, temperature, and context-to-component weights are researcher-specified parts of the hypothetical synthetic operationalization. The compiler verifies that every cited source identifier is registered and that the contract prohibits clinical use. It then produces a nine-component diagonal Gaussian mixture
\begin{equation}
p_{\KTCM}(Z\mid C=c)=\sum_{k=1}^{9}\omega_{ck}
\Normal\!\left(Z;\bm{a}_k,\operatorname{diag}(\bm{v}_k)\right).
\label{eq:knowledge-contextual-prior}
\end{equation}
Component locations $\bm{a}_k$ are deterministic functions of the versioned synthetic pattern rules; component variances, context cues, and the softmax temperature are declared in the contract. The three contexts are operationalized from synthetic intake metadata. Terminology provenance supports auditability of the representation, not clinical or biomedical validity of the asserted context-to-pattern weights.

The synthetic renderer reads only a separately frozen generator-truth file. It does not import the knowledge directory, prior builder, or compiled parameter object. Conversely, the knowledge compiler does not read generator truth, rendered cases, latent targets, or outcomes. This software separation prevents direct parameter sharing. It does not make the two constructions historically independent: their semantic alignment is deliberately designed so that the synthetic experiment can test an informative but imperfect prior. Every point artifact records the hashes of the generator specification, prior inputs, configuration, training code, and checkpoints.

For encoders learned under a marginal Gaussian reference $p_0(Z)$, the contextual fusion rule is
\begin{equation}
q_{\KTCM}(Z\mid X^{(\mathcal A)},C)
\propto p_{\KTCM}(Z\mid C)\,p_0(Z)^{-|\mathcal A|}
\prod_{m\in\mathcal A}q_m(Z\mid X^{(m)}).
\label{eq:contextual-committee}
\end{equation}
Equation~\ref{eq:contextual-committee} first removes the reference-prior contribution from every conditional encoder and then inserts the declared contextual prior. When $p_{\KTCM}=p_0$, it reduces to Equation~\ref{eq:bcm}. For each mixture component, Gaussian natural parameters are combined analytically; normalized component masses are then moment-matched to one diagonal Gaussian for the frozen downstream bridge. Nonpositive component precision is reported, and a declared $10^{-4}$ floor keeps evaluation numerically defined. This is an approximate conditional-expert calculation under common-reference-prior, conditional-independence, and calibration assumptions; it is not exact Bayesian inference in the neural-network parameters.

\subsection{Encoder-Level Prior Regularization}

Inference-time contextual fusion is distinct from the encoder-level training regularizer. For a prior condition $r$, each modality encoder may be trained with
\begin{equation}
\mathcal{L}_{m}^{(r)}=\mathcal{L}_{m}
+\beta\,\frac{1}{d}\KL\!\left[q_m(Z\mid X^{(m)})\,\|\,p_r(Z\mid C)\right],
\qquad d=4,
\label{eq:stage0-prior-loss}
\end{equation}
using four reparameterized samples and fixed $\beta\in\{0.005,0.05\}$. The matched conditions are: no structured regularization; the context-marginalized knowledge prior; aligned privileged context; and a fixed wrong permutation of the same context weights. The privileged conditions expose context only to the regularizer, never to a modality encoder. Initialization, cases, minibatch order, optimizer settings, bridge training, and evaluation cohorts are paired within each seed. Equation~\ref{eq:stage0-prior-loss} is distributional regularization, not a Bayesian posterior over network weights.

\subsection{Prior Corruption and Conflict Audit}

Prior misspecification is evaluated at inference without retraining the shared encoders. Context-weight contamination interpolates between aligned weights and a fixed permutation at fractions 0, 0.25, 0.50, 0.75, and 1. Structural challenges include small, medium, and large translations; 15- and 45-degree rotations; an axis swap; narrow and broad variance scaling; flattened and sharpened weights; and missing, merged, and spurious components. Context-feature noise is applied equally to the knowledge, learned-context, and late-concatenation systems.

The conflict score uses only predicted evidence and declared context. We report AUROC, average precision, the wrong-context flag rate at an aligned 95th-percentile threshold, and association between score increase and excess prediction error. It does not use the true latent target and is not a prior-predictive Bayesian test. A useful warning signal must discriminate aligned from corrupted context and predict damage; detecting an intentionally altered identifier alone would not satisfy that requirement. No corruption result can establish that a TCM theory is true or false beyond this operationalization.

\subsection{Continuous Downstream Bridge}

The primary encoder target $Z\in\mathbb{R}^4$ is separated from a continuous downstream vector $Y\in[-1,1]^5$. A small bridge receives the fused mean and log-variance, keeps the modality encoders frozen, and returns
\begin{equation}
q(Y\mid\mu_{\mathrm{P}},\sigma_{\mathrm{P}}^2)
=\Normal\!\left(\mu_Y,\operatorname{diag}(\sigma_Y^2)\right).
\label{eq:bridge}
\end{equation}
It is trained by Gaussian NLL against the restricted continuous targets. Three-valued anchors derived by thresholding $Y$ are secondary summaries only. The model is therefore not a five-way clinical classifier and does not output diagnoses or treatment recommendations.

\subsection{Uncertainty, Identifiability, and Falsifiability}

Reported variance combines aleatoric and model-dependent effects and is not guaranteed to be calibrated under generator or real-world shift. We therefore report NLL and empirical coverage at 50\%, 80\%, and 95\%. Latent coordinates also remain subject to construct and identifiability limitations: even a recoverable synthetic coordinate is meaningful only under its generating convention. Structural corruptions test sensitivity to this declared operationalization, not whether a medical theory is true. Stronger future tests must freeze knowledge before independently selecting multiple generator families or real datasets, evaluate repeated measurements, and predict outcomes defined independently of the concepts used to construct the prior.

\section{Longitudinal and Decision Scope}
\label{sec:longitudinal-decision}

The framework extends the static state with
\begin{equation}
p(Z_{t+1}\mid Z_t,\Delta t_t,A_t,C_t,\KTCM),
\label{eq:transition}
\end{equation}
where $\Delta t_t$ is elapsed time, $A_t$ an observed intervention or management action, and $C_t$ contextual covariates. Filtering estimates $p(Z_t\mid X_{1:t})$, smoothing revises earlier states using later observations, and forecasting propagates uncertainty to future visits. Conditioning on $A_t$ is not equivalent to estimating a causal effect under $\mathrm{do}(A_t)$. Causal treatment-effect estimation would require a separate identification argument, confounder specification, positivity assumptions, and an appropriate design.

The decision layer is also deliberately downstream. Given a posterior $q(Z_t)$, a constrained system might abstain, request another modality, repeat an unreliable measurement, or present several hypotheses. An LLM may explain a fixed numerical record and its provenance, but it cannot silently replace the posterior or authorize treatment. The present experiment contains neither trajectories nor decisions; this section states only the intended separation of responsibilities.

\section{Synthetic Experiment}
\label{sec:experiment}

\subsection{Scope and Independent Construction}

The experiment evaluates a static synthetic neural-evidence workflow under controlled source separation. The renderer consumes only \texttt{generator\_truth\_v2.json}; the knowledge compiler consumes only versioned knowledge files. Both produce nine-component, four-dimensional structures with three contexts, but they use different parameter objects and code paths. The generator has unequal component locations, variances, context weights, modality renderers, and a separately declared nonlinear five-dimensional outcome mechanism. The study therefore asks whether a deliberately related but nonidentical knowledge construction helps finite-sample learning, not whether the prior rediscovers parameters that were passed directly from the renderer.

For every training size and seed, the generator creates a training cohort and a disjoint 600-case development-validation cohort. Each case contains four aligned observations: a tongue-like RGB image, a pulse-like waveform, a 20-field structured inquiry with observation mask, and a locally generated byte sequence, together with three observed context features. Restricted four-dimensional latent targets and continuous five-dimensional outcomes are available only to the training and evaluation paths. Generator metadata, component identity, file paths, and split membership are never model inputs.

The scientific confirmation cohorts are determined by a prespecified seed offset and are inaccessible during development. The confirmation command writes a permanent create-once access ledger before generating any confirmation sample and refuses retries, including after interruption. Engineering rehearsals use a distinct seed family and output root. This is stronger leakage control than calling an already inspected split ``sealed.''

\subsection{Implemented Models}

Table~\ref{tab:implemented-models} lists the four independent encoders. All heads emit four means and strictly positive variances with minimum variance $10^{-4}$. The inquiry encoder concatenates masked coded values with the observation mask. The byte-level text representation avoids fitting a vocabulary on validation data.

\begin{table}[H]
\centering
\small
\begin{tabularx}{\textwidth}{@{}p{0.13\textwidth}p{0.20\textwidth}X p{0.20\textwidth}@{}}
\toprule
Modality & Input & Encoder & Evidence output \\
\midrule
Tongue & RGB image & Three strided 2-D convolutions and global pooling & $4$ means, $4$ variances \\
Pulse & Length-256 waveform & Residual 1-D CNN and global pooling & $4$ means, $4$ variances \\
Inquiry & 20 codes plus mask & Two-layer masked MLP, hidden width 64 & $4$ means, $4$ variances \\
Text & Up to 384 UTF-8 bytes & 32-D byte embedding; kernels 3, 5, 7; max pooling & $4$ means, $4$ variances \\
\bottomrule
\end{tabularx}
\caption{Implemented modality encoders. Descriptions follow the executable model definitions; they are deliberately compact and CPU-sized.}
\label{tab:implemented-models}
\end{table}

Each encoder is optimized independently by Equation~\ref{eq:encoder-nll}, or by Equation~\ref{eq:stage0-prior-loss} in the prior-regularized conditions. Selected encoders are frozen before bridge training. The primary shared-encoder comparison contains four systems: marginal Gaussian committee fusion; the source-registered, researcher-specified contextual mixture in Equation~\ref{eq:contextual-committee}; an ordinary neural predictor that concatenates the same three context features to the marginal fused mean and log-variance; and a strong late-concatenation predictor that receives every expert mean and log-variance plus the same context features. Thus a contextual-prior advantage cannot be attributed merely to access to context. Frozen encoder evidence is reused within a point; this changes computation time but not model inputs or objectives.

\subsection{Preregistered Training and Evaluation Protocol}

The development grid contains four nested training sizes, 600, 1,200, 2,400, and 4,800, and ten paired seeds, 917302--917311, for 40 model points. Every point uses 30 maximum epochs, batch size 32, patience 8, four Monte Carlo samples for encoder-level KL regularization, fixed $\beta\in\{0.005,0.05\}$, 600 development-validation cases, and a two-thread CPU limit under the safe default scheduler. Checkpoint selection uses development Gaussian NLL. Shared source files, configuration, generator truth, knowledge inputs, checkpoints, and final result JSON are hashed; point and grid locks prevent duplicate writers; JSON outputs are atomic.

Primary continuous metrics are MAE, RMSE, Gaussian NLL, and empirical 50\%, 80\%, and 95\% coverage. Derived five-axis anchor accuracy and exact match are secondary summaries. Prespecified paired MAE contrasts compare contextual knowledge with marginal committee fusion, learned context, and late concatenation; fully wrong context with aligned knowledge; and each Stage-0 condition with its matched control. Aggregation uses all ten paired seeds at every size and reports bootstrap 95\% intervals. Failure to exclude zero is not interpreted as equivalence.

After all 40 development artifacts passed schema, completeness, seed, hash, and code-provenance validation, the aggregate and analysis specification were frozen. The explicit authorization token then ran the create-once confirmation evaluation. Ten unique 600-case confirmation cohorts evaluated the 40 frozen training points without retraining or model selection. The claim-decision script refuses development or engineering-smoke aggregates and applies a conservative rule: a contextual benefit must be directionally supported at both low-data sizes; a specifically Bayesian interpretation additionally requires separation from at least one equal-information neural control; and wrong-prior harm is a separate claim.

The experiment also reports graded context contamination, structural corruption, context-feature noise, fusion precision diagnostics, and conflict discrimination. These are synthetic stress tests, not evidence of robustness to real acquisition devices, patients, institutions, or language shift.

\section{Results}
\label{sec:results}

\subsection{Frozen Confirmation and Primary Comparison}

\begin{figure}[H]
\centering
\includegraphics[width=\textwidth]{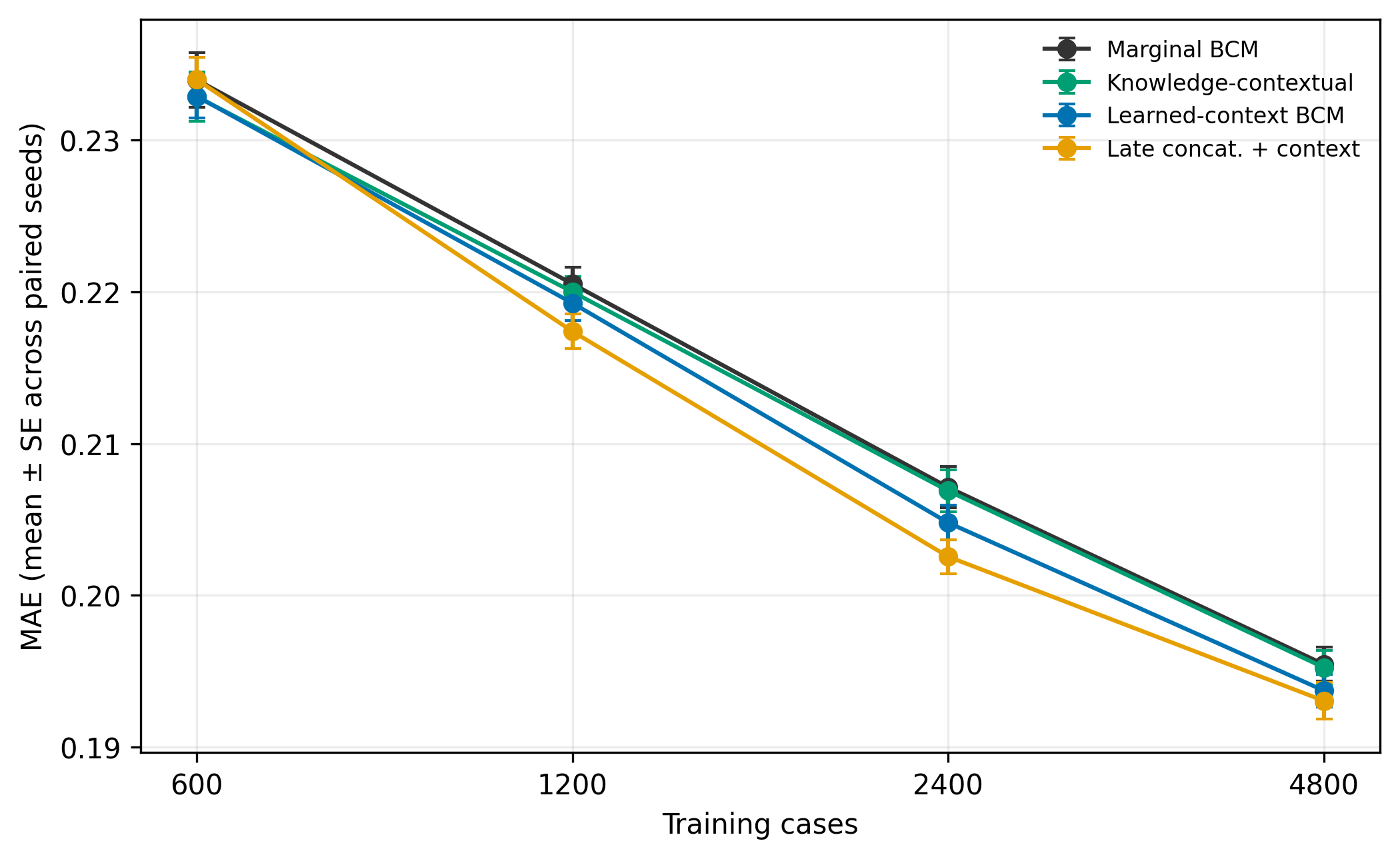}
\caption{Confirmation learning curves across ten paired seeds. Absolute MAE decreases with training size. The researcher-specified synthetic contextual prior has a small low-data advantage over marginal BCM, whereas the equal-information learned and late-context systems become stronger as sample size increases.}
\label{fig:v7-learning-curves}
\end{figure}

The evidence gate completed without retraining or post-freeze selection: all ten confirmation cohorts and all 40 frozen training points passed validation. The full SHA-256 digests of the frozen specification, confirmation aggregate, and claim-decision record are embedded in the manuscript source. Table~\ref{tab:v7-primary-mae} and Figure~\ref{fig:v7-learning-curves} report confirmation-only estimates; values after $\pm$ are sample standard deviations across the ten paired seeds.

\begin{table}[H]
\centering
\small
\begin{tabular}{@{}rcccc@{}}
\toprule
Train $n$ & Marginal BCM & Knowledge context & Learned context & Late context \\
\midrule
600  & $0.2340\pm0.0057$ & $0.2329\pm0.0052$ & $0.2329\pm0.0045$ & $0.2340\pm0.0046$ \\
1,200 & $0.2205\pm0.0034$ & $0.2200\pm0.0033$ & $0.2193\pm0.0037$ & $0.2174\pm0.0036$ \\
2,400 & $0.2071\pm0.0043$ & $0.2069\pm0.0044$ & $0.2048\pm0.0036$ & $0.2026\pm0.0036$ \\
4,800 & $0.1955\pm0.0036$ & $0.1952\pm0.0035$ & $0.1937\pm0.0034$ & $0.1931\pm0.0038$ \\
\bottomrule
\end{tabular}
\caption{Confirmation MAE for the four shared-encoder systems. Lower is better. Context features are matched across the knowledge, learned-context, and late-context systems.}
\label{tab:v7-primary-mae}
\end{table}

The paired knowledge-minus-marginal MAE was $-0.001106$ (bootstrap 95\% interval $[-0.002192,-0.000318]$) at 600 cases and $-0.000528$ ($[-0.000876,-0.000232]$) at 1,200, satisfying the prespecified low-data contextual-benefit rule. At 2,400 and 4,800 the intervals included zero. The Bayesian-specific rule was not satisfied: knowledge context was indistinguishable from the learned and late controls at 600, then worse than learned context by $0.000739$, $0.002101$, and $0.001523$ at 1,200, 2,400, and 4,800, respectively; it was also worse than late context by $0.002606$, $0.004340$, and $0.002198$. The integrated log-size learning-curve areas were 0.214123 (marginal), 0.213651 (knowledge), 0.212454 (learned context), and 0.211157 (late context), with lower values better. These results support contextual information over marginal fusion at low data, but do not isolate a specifically Bayesian performance advantage.

\begin{figure}[H]
\centering
\includegraphics[width=\textwidth]{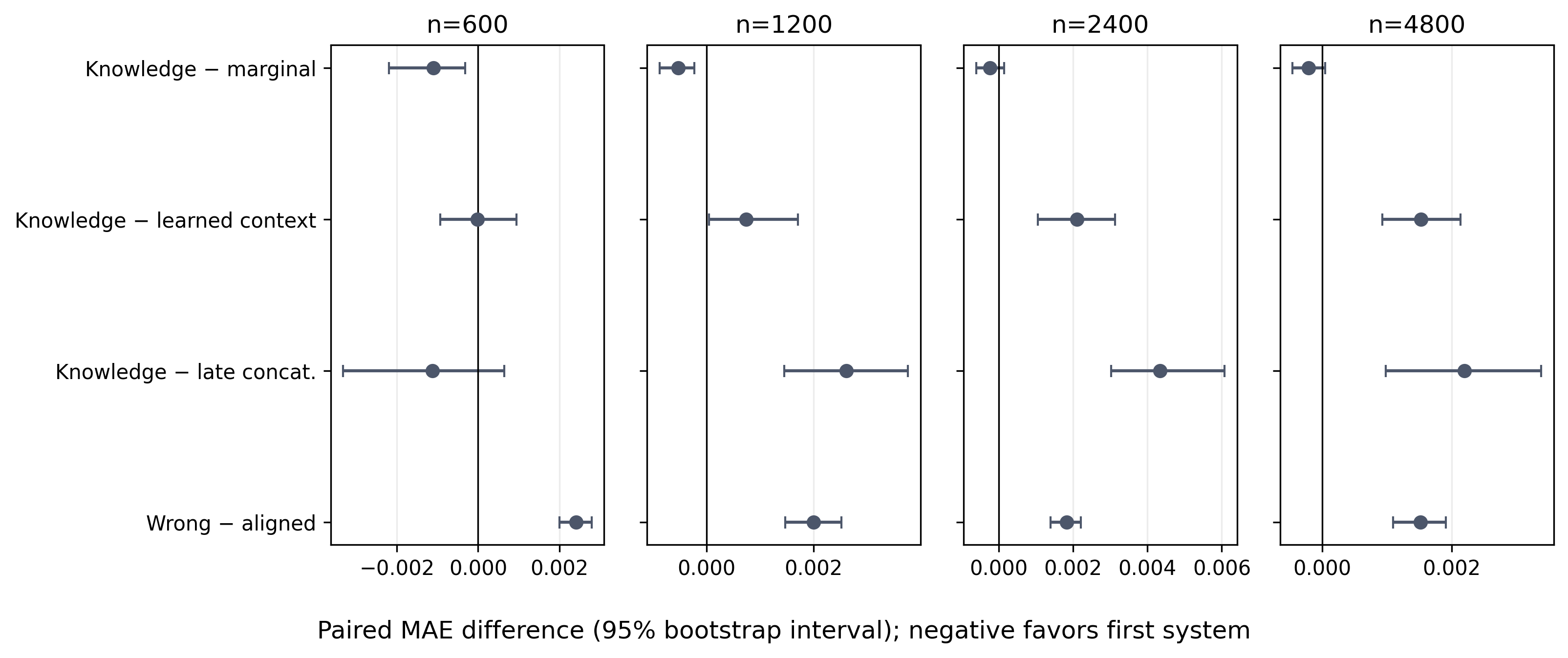}
\caption{Prespecified paired MAE contrasts with bootstrap 95\% intervals. Negative values favor knowledge context for the first three contrasts; positive values in the wrong-minus-aligned contrast indicate harm. An interval crossing zero is reported as inconclusive, not equivalent.}
\label{fig:v7-paired-effects}
\end{figure}

\subsection{Uncertainty and Encoder-Level Prior Regularization}

Table~\ref{tab:v7-uncertainty} shows that the knowledge-context system's interval calibration improved with sample size. Coverage was below nominal at the two smaller sizes, especially for the nominal 80\% interval, and approached nominal values at 2,400--4,800. Bayesian notation therefore did not remove the need to evaluate calibration.

\begin{table}[H]
\centering
\small
\begin{tabular}{@{}rccccc@{}}
\toprule
Train $n$ & MAE & RMSE & Gaussian NLL & 80\% cov. & 95\% cov. \\
\midrule
600 & $0.2329\pm0.0052$ & $0.2942\pm0.0047$ & $0.2049\pm0.0283$ & $0.760\pm0.018$ & $0.922\pm0.012$ \\
1,200 & $0.2200\pm0.0033$ & $0.2798\pm0.0030$ & $0.1167\pm0.0232$ & $0.783\pm0.012$ & $0.936\pm0.006$ \\
2,400 & $0.2069\pm0.0044$ & $0.2663\pm0.0038$ & $0.0191\pm0.0223$ & $0.797\pm0.014$ & $0.949\pm0.006$ \\
4,800 & $0.1952\pm0.0035$ & $0.2540\pm0.0037$ & $-0.0579\pm0.0166$ & $0.811\pm0.007$ & $0.954\pm0.005$ \\
\bottomrule
\end{tabular}
\caption{Confirmation point error, Gaussian score, and empirical interval coverage for knowledge-context fusion. Values are mean $\pm$ sample standard deviation across seeds.}
\label{tab:v7-uncertainty}
\end{table}

\begin{figure}[H]
\centering
\includegraphics[width=\textwidth]{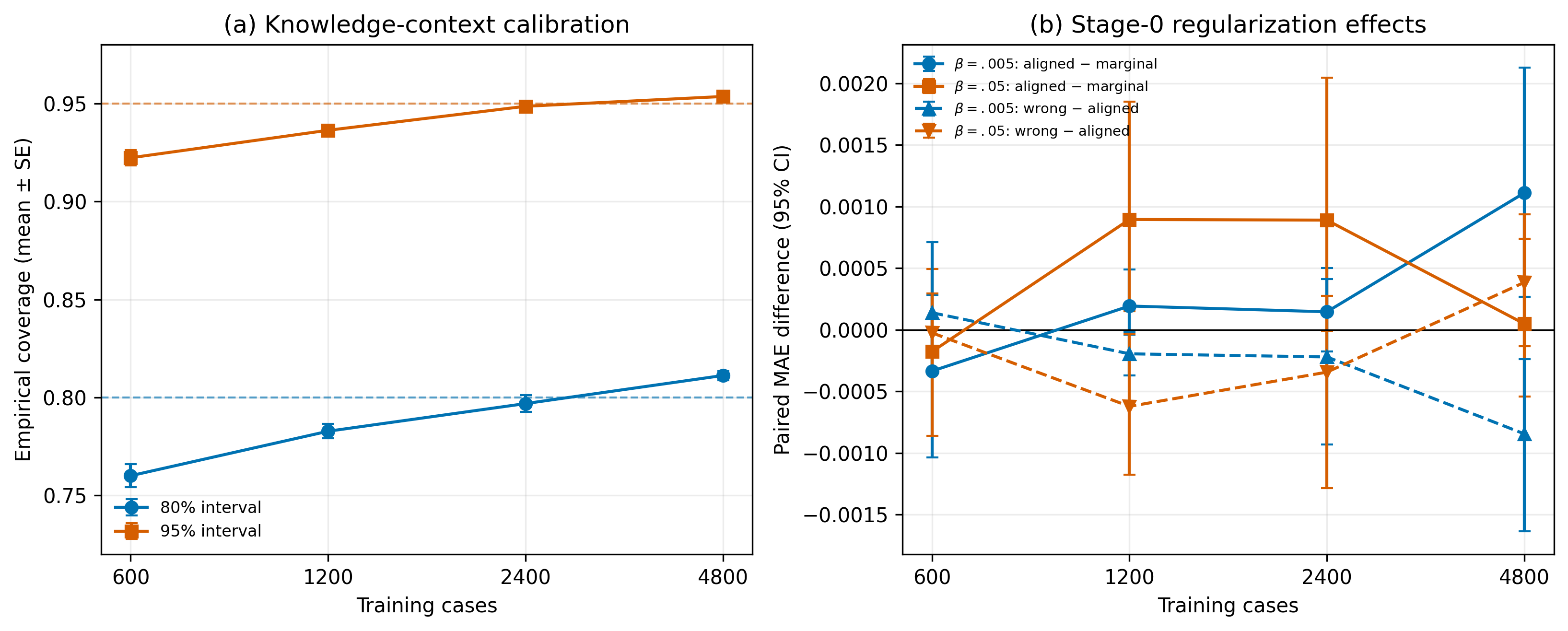}
\caption{Additional confirmation diagnostics. (a) Empirical coverage of the knowledge-context system across training sizes; error bars are standard errors over ten paired seeds and dashed lines mark nominal 80\% and 95\% coverage. (b) Prespecified encoder-level regularization contrasts with bootstrap 95\% intervals. Each legend label is written as the first system minus the second, so negative aligned-minus-marginal values favor aligned regularization and positive wrong-minus-aligned values indicate harm from the wrong regularizer.}
\label{fig:v7-uncertainty-stage0}
\end{figure}

Encoder-level distributional regularization did not provide the expected aligned-prior learning advantage. Most aligned-minus-marginal and wrong-minus-aligned intervals crossed zero. Moreover, aligned regularization was worse than marginal BCM at 1,200 cases for $\beta=0.05$ ($+0.000896$, interval $[0.000150,0.001851]$) and at 4,800 for $\beta=0.005$ ($+0.001111$, $[0.000266,0.002127]$). Several wrong-minus-aligned contrasts unexpectedly favored the wrong regularizer. The magnitudes were small, but the direction reversals mean that this implementation cannot support the hypothesis that an aligned prior accelerates learning while a wrong prior delays it. Figure~\ref{fig:v7-uncertainty-stage0} places this mixed regularization result beside the distinct finding that predictive coverage improves with sample size.

\subsection{Misspecification and Conflict Feedback}

\begin{figure}[H]
\centering
\includegraphics[width=\textwidth]{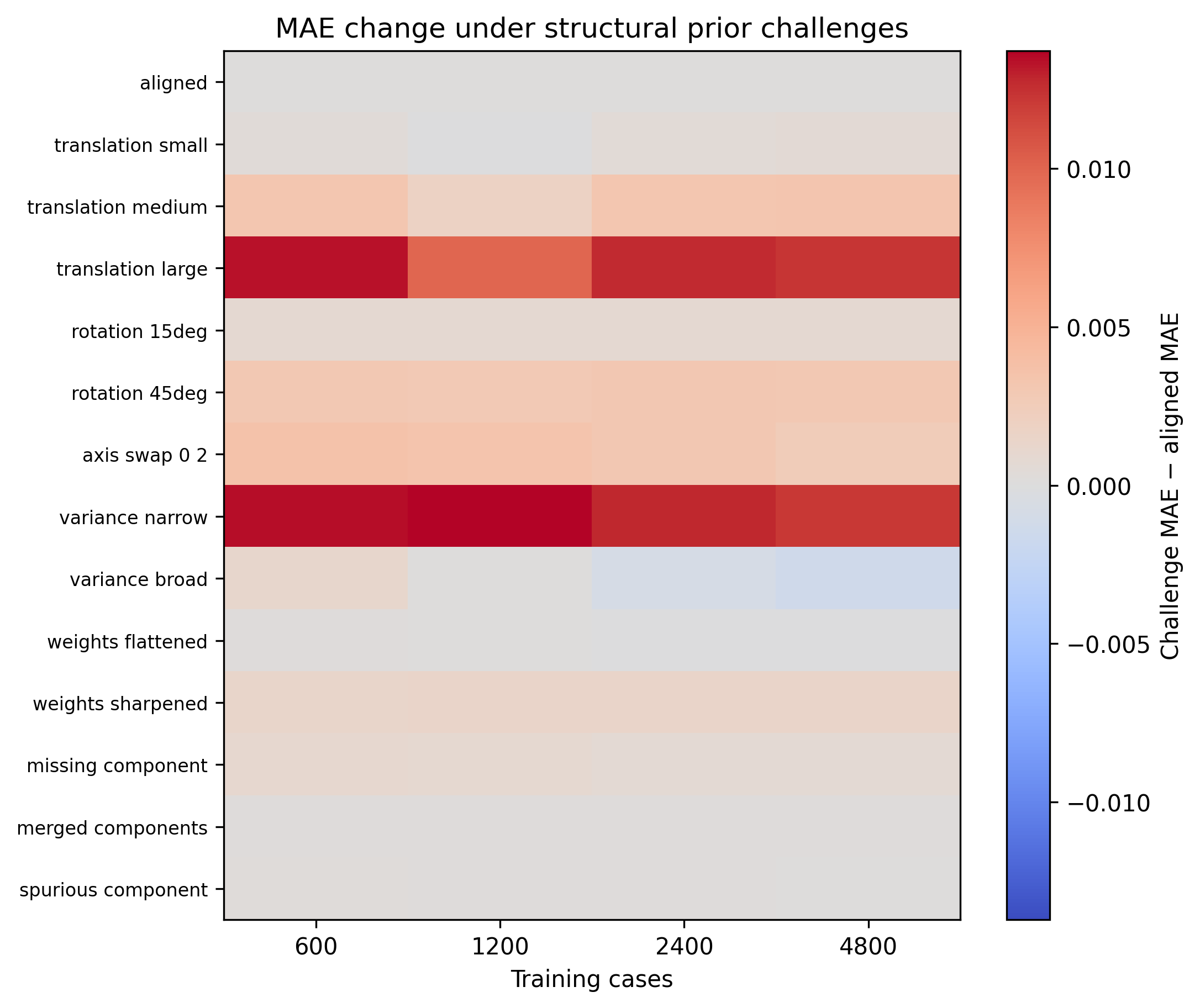}
\caption{Confirmation audit of graded context contamination and structurally distinct prior errors. Effects are specific to the declared four-dimensional synthetic latent convention and do not test the validity of TCM theory.}
\label{fig:v7-structural}
\end{figure}

Fully wrong inference-time context increased MAE at every size (Table~\ref{tab:v7-conflict}). Weight contamination produced an approximately graded response: aligned-to-fully-wrong MAE rose from 0.232871 to 0.235281 at 600, from 0.220003 to 0.222002 at 1,200, from 0.206891 to 0.208714 at 2,400, and from 0.195248 to 0.196766 at 4,800. Figure~\ref{fig:v7-structural} shows that narrow variance and large translation were the most damaging structural errors, adding roughly 0.012--0.014 MAE. Axis swapping and 45-degree rotation caused smaller penalties around 0.0025--0.0036, while weight flattening, merged components, and a spurious component were nearly neutral under this generator. The audit therefore resolves a severity spectrum rather than only an extreme translated prior.

\begin{table}[H]
\centering
\small
\begin{tabular}{@{}rccccc@{}}
\toprule
Train $n$ & Wrong excess MAE [95\% CI] & AUROC & AP & Wrong flag rate & Damage $\rho$ \\
\midrule
600 & $0.002410\ [0.002001,0.002795]$ & 0.620 & 0.608 & 0.107 & 0.014 \\
1,200 & $0.001999\ [0.001470,0.002514]$ & 0.614 & 0.606 & 0.112 & 0.022 \\
2,400 & $0.001823\ [0.001390,0.002213]$ & 0.612 & 0.603 & 0.109 & 0.031 \\
4,800 & $0.001519\ [0.001096,0.001913]$ & 0.610 & 0.599 & 0.105 & 0.045 \\
\bottomrule
\end{tabular}
\caption{Wrong-context harm and descriptive conflict diagnostics. The threshold is the aligned-score 95th percentile; $\rho$ is the seed-averaged Spearman association between score increase and case-level prediction damage.}
\label{tab:v7-conflict}
\end{table}

Despite reliable aggregate harm, conflict discrimination was weak (AUROC 0.610--0.620; average precision 0.599--0.608), and the wrong-context flag rate was only 0.105--0.112 at the aligned 95th-percentile threshold. More importantly, the association with case-level damage was near zero ($\rho=0.014$--$0.045$). The score can weakly distinguish constructed context conditions but cannot yet provide useful case-level feedback. All results remain conditional on one synthetic generator family and its declared context mechanism; none supports a clinical, diagnostic, treatment, causal, longitudinal, or transfer inference.
\section{Discussion}
\label{sec:discussion}

\subsection{Nine-Layer Architecture as a System Contribution}

The first contribution of Alpha-TCM is independent of whether the prototype prior improves MAE. The architecture establishes an information and authority order for computational TCM: semantic assertions precede acquisition; acquisition is qualified by availability, validity, and quality; structured knowledge remains distinguishable from case evidence; inference precedes longitudinal updating and decisions; and language generation remains subordinate to numerical records and human governance. This decomposition addresses a systems problem that an end-to-end classifier cannot resolve by scale alone. Its present value is conceptual precision, traceability, and failure containment, not demonstrated superiority over existing lifecycle, reporting, or governance frameworks and not empirical validation of unevaluated layers.

The architecture also creates explicit falsification obligations. A prior can be removed or corrupted; a modality can be missing without becoming negative evidence; uncertainty must survive into downstream interfaces; and longitudinal or treatment language cannot inherit causal authority from a static predictor. These contracts make future implementations comparable because each layer has declared inputs, outputs, failure modes, and validation requirements.

\subsection{Prototype Audit of the Knowledge--Evidence Interface}

The second contribution is the static prototype that instantiates Layers 1--5 and subjects their central interface to controlled evaluation. The experiment asks whether a provenance-bearing TCM operationalization, compiled without reading the generator implementation, supplies useful context-dependent inductive structure beyond marginal fusion and beyond neural systems given the same context. The prototype therefore provides empirical evidence about one architecture pathway while preserving a meaningful failure outcome.

The frozen confirmation selected the second prespecified interpretation: knowledge contextual fusion improved over marginal BCM at both low-data sizes, but did not separate from either equal-information neural control at 600 and was worse than both from 1,200 onward. This is a mixed audit finding: it supports useful context-dependent inductive structure in the controlled low-data regime, not the stronger proposition that the Bayesian representation is predictively superior. This distinction is central rather than cosmetic: the experiment demonstrates that the framework's knowledge interface is challengeable and that access to context, representation choice, and probabilistic semantics must be evaluated separately.

\subsection{Bayesian and Conflict-Feedback Interpretation}

The Bayesian role is explicit but conditional. Equation~\ref{eq:bcm} corrects repeated reference-prior information in discriminative conditional experts, and Equation~\ref{eq:contextual-committee} replaces that reference with a context-dependent mixture. Encoder-level prior regularization separately asks whether privileged structure shapes finite-sample representation learning. These calculations rely on approximate calibration, a common reference prior, conditional independence, and tractable Gaussian products. Equal-information controls test predictive specificity; they do not prove the factorization assumptions.

The conflict audit is deliberately separate. Wrong context caused reproducible aggregate harm, but AUROC remained near 0.61 and damage correlation near zero. The present score therefore distinguishes experimental conditions only weakly and cannot guide case-level trust. The result is a negative finding for the current feedback mechanism, not evidence that prior conflict is undetectable in principle. No result allows the model to adjudicate the general validity of TCM theory.

\subsection{Boundary of the Implemented Claim}

The implementation demonstrates a traceable Layers 1--5 pathway, heteroscedastic conditional encoders, corrected marginal committee fusion, contextual mixture prior replacement, distribution-level regularization, equal-information controls, and a reproducible synthetic audit. It does not validate the complete nine-layer system, establish posterior uncertainty over network parameters, model practitioners or longitudinal dynamics, demonstrate decision utility, or provide a clinical prior-predictive conflict test. Calling it a validated digital human, diagnostic system, or treatment-support tool would be unsupported.

\section{Limitations and Validation Roadmap}
\label{sec:limitations}

The two contribution tracks have different limitations. The nine-layer architecture is a formal system design, but Layers 6--9 and most Layer-3 quality functions are not implemented or empirically evaluated. It has not undergone expert consensus, user study, comparative fault injection, or formal verification against alternative system decompositions. Completeness of a contract does not establish necessity, usability, safety, or clinical validity. The prototype covers only a static Layers 1--5 pathway and uses no real patient data; it therefore cannot establish diagnostic accuracy, clinical calibration, validity of TCM constructs, longitudinal forecasting, treatment effects, workflow utility, or external transfer. Synthetic latent recovery may exploit renderer regularities that do not occur in practice.

Second, software separation is not semantic independence. The knowledge compiler and generator use distinct files, code paths, and parameter objects, but their broad context--pattern alignment was designed by the same research project. The experiment tests whether this versioned operationalization is useful under controlled mismatch; it does not show that historical sources independently predicted an unseen data-generating process. A stronger study would freeze the knowledge representation before an independently authored generator or real dataset is selected.

Third, only one four-dimensional generator family, one outcome mechanism, three synthetic contexts, and compact CPU-sized architectures are evaluated. The downstream target is generated within the same overall simulator even though its mechanism is separately declared. Generator-family shift, unrelated outcomes, invalid-domain inputs, acquisition-device changes, text paraphrase families, and alternative model capacities remain untested.

Fourth, contextual committee fusion assumes calibrated conditional experts, a common marginal reference prior, and conditional independence. The reference is a training-target-fitted diagonal Gaussian; contextual components are diagonal; and the final mixture is moment-matched for the bridge. Precision flooring makes failure measurable but does not repair a false factorization. Correlated-modality and calibration sweeps are needed before generalizing the fusion behavior.

Fifth, the structural corruptions are controlled interventions on a synthetic prior. Their severity is interpretable only relative to the chosen latent convention. The conflict score is observation-side and target-free, but it is not a prior-predictive Bayesian test. Ten paired seeds improve uncertainty assessment over the earlier three-seed study, yet they do not substitute for multiple generator families or external data.

Sixth, the source registry establishes provenance rather than evidential truth. The contextual-prior contract is explicitly marked a hypothetical synthetic operationalization and prohibits clinical use. A predictive improvement would show that the encoded relation helps this task, not that a TCM construct is biologically real; a wrong-prior penalty would falsify only the implemented mapping.

A staged validation sequence prevents stronger claims from outrunning evidence:
\begin{enumerate}
  \item freeze and audit a small source-grounded semantic scope before dataset selection;
  \item test multiple independently authored synthetic generators and unrelated outcome mechanisms;
  \item measure repeatability and acquisition quality for each real modality;
  \item validate static prediction and calibration with patient-level splits and independent targets;
  \item collect overlapping practitioner assessments to estimate disagreement rather than erase it;
  \item compare generic, random, fixed domain, learned-context, and adaptive priors under matched information and capacity;
  \item evaluate repeated-visit forecasting at prespecified horizons; and
  \item only then study human-supervised decision utility, causal questions, and safety prospectively.
\end{enumerate}

\section{Conclusion}

Alpha-TCM makes two co-primary contributions. It first establishes a nine-layer TCM-oriented system architecture that allocates meaning, evidence, uncertainty, temporal state, decision authority, language interaction, and human governance to explicit interfaces. It then instantiates the static Layers 1--5 pathway in a controlled prototype that makes the architecture's knowledge--evidence contract empirically testable.

The prototype compiles a source-registered, researcher-specified synthetic contextual prior from versioned knowledge files, keeps the multimodal renderer in a separate implementation path, removes duplicated marginal-prior information before introducing $p_{\KTCM}(Z\mid C)$, and compares the result with equal-information neural controls and deliberately wrong priors. Frozen confirmation supports a small contextual benefit over marginal fusion at 600 and 1,200 cases, but it does not isolate a specifically Bayesian predictive advantage. Wrong inference-time context causes graded aggregate harm, encoder-level prior regularization does not reliably improve learning, and the present conflict score does not identify case-level damage.

These mixed results strengthen the paper's main methodological boundary: a source-registered, TCM-motivated synthetic structure can be represented, versioned, inserted, removed, corrupted, and empirically challenged without being treated as unquestioned truth. The nine-layer architecture and its prototype audit remain co-primary contributions, but neither constitutes validation or invalidation of TCM theory, and neither supports clinical diagnosis, treatment, causality, longitudinal performance, or real-world transfer.

\end{document}